\documentclass{article}

\usepackage[final, nonatbib]{nips_2016}

\usepackage[utf8]{inputenc} % allow utf-8 input
\usepackage[T1]{fontenc}    % use 8-bit T1 fonts
\usepackage{hyperref}       % hyperlinks
\usepackage{url}            % simple URL typesetting
\usepackage{amsfonts}       % blackboard math symbols
\usepackage{nicefrac}       % compact symbols for 1/2, etc.
\usepackage{microtype}      % microtypography
\usepackage{graphicx}
\usepackage{xcolor}
\usepackage{amsmath}
\usepackage{hyphenat}
\usepackage{bbm}
\usepackage{bm}
\usepackage{amssymb}
\usepackage{wrapfig}
\usepackage{multirow}
\usepackage{mathtools}
\usepackage{esint}
\usepackage{mathtools}
\usepackage[font=small,labelfont=bf]{caption}
\usepackage[neverdecrease]{paralist}
\DeclareMathSizes{10}{9}{6}{5}

\definecolor{sig001}{HTML}{0000FF}
\definecolor{sig01}{HTML}{008000}
\definecolor{sig05}{HTML}{FF0000}

\usepackage{booktabs}       % professional-quality tables
\usepackage{environ}
\NewEnviron{Equation}[1]{%
\begin{equation}
  \scalebox{#1}{$\BODY$}
\end{equation}
}
\NewEnviron{Equation*}[1]{%
\begin{equation*}
  \scalebox{#1}{$\BODY$}
\end{equation*}
}

\title{Beyond Exchangeability: The Chinese Voting Process}

\author{
  Moontae Lee\\
  Dept. of Computer Science\\
  Cornell University\\
  Ithaca, NY 14853 \\
  \texttt{moontae@cs.cornell.edu} \\
  \And
  Seok Hyun Jin\\
  Dept. of Computer Science\\
  Cornell University\\
  Ithaca, NY 14853 \\
  \texttt{sj372@cornell.edu} \\
  \And
  David Mimno\\
  Dept. of Information Science\\
  Cornell University\\
  Ithaca, NY 14853 \\
  \texttt{mimno@cornell.edu} \\
}

\begin{document}

\newcommand{\E}{\mathbb{E}}

\newcommand{\M}[3]{\smash{{#1}_{\mkern-3mu #2}^{\mkern-2mu (#3)}}}
\newcommand{\MU}[2]{\smash{{#1}^{\mkern-1mu (#2)}}}
\newcommand{\MA}[3]{\smash{{#1}_{\mkern-3mu #2}^{\mkern-2mu #3}}}
\newcommand{\MM}[3]{{#1}_{\mkern-3mu #2}^{\mkern-1mu (#3)}}
\newcommand{\MP}[3]{{#1}_{\mkern-3mu #2}^{\mkern-6mu +(#3)}}
\newcommand{\MN}[3]{{#1}_{\mkern-3mu #2}^{\mkern-6mu -(#3)}}

\newcommand{\DM}[1]{\textcolor{red}{[#1]}}
\newcommand{\ML}[1]{\textcolor{blue}{[#1]}}
\newcommand{\SJ}[1]{\textcolor{green}{[#1]}}

\maketitle

%%%%%%%%%%%%%%%%%%%%%%%%%%%%%%%%%%%%%%%%%%%%%%%%%%%%%%%%%%%%%%%%%%
\begin{abstract}
  %%%%%%%%%%%%
% Abstract %
%%%%%%%%%%%%

Many online communities present user\hyp{}contributed responses such as reviews of products and answers to questions.
User-provided helpfulness votes can highlight the most useful responses, but voting is a social process that can gain momentum based on the popularity of responses and the polarity of existing votes.
We propose the Chinese Voting Process (CVP) which models the evolution of helpfulness votes as a self-reinforcing process dependent on position and presentation biases.
We evaluate this model on Amazon product reviews and more than 80 StackExchange forums, measuring the intrinsic quality of individual responses and behavioral coefficients of different communities.

\end{abstract}

%%%%%%%%%%%%%%%%%%%%%%%%%%%%%%%%%%%%%%%%%%%%%%%%%%%%%%%%%%%%%%%%%%
\section{Introduction}
  %%%%%%%%%%%%%%%%%%%
% 1. Introduction %
%%%%%%%%%%%%%%%%%%%
\label{sec:introduction}

With the expansion of online social platforms, user\hyp{}generated content has become increasingly influential.
Customer reviews in e\hyp{}commerce like Amazon are often more helpful than editorial reviews \cite{KRC:2012}, and question answers in Q\&A forums such as StackOverflow and MathOverflow are highly useful for coders and researchers \cite{Mamykina:2011,Tausczik:2014}.
Due to the diversity and abundance of user content, promoting better access to more useful information is critical for both users and service providers.
Helpfulness voting is a powerful means to evaluate the quality of user responses (i.e., reviews\slash{}answers) by the wisdom of crowds.
While these votes are generally valuable in aggregate, estimating the true quality of the responses is difficult because users are heavily influenced by previous votes.
We propose a new model that is capable of learning the intrinsic quality of responses by considering their social contexts and momentum.

Previous work in self\hyp{}reinforcing social behaviors shows that although inherent quality is an important factor in overall ranking, users are susceptible to \textit{position bias} \cite{Salganik:2006,Salganik:2008}.
Displaying items in an order affects users: top-ranked items get more popularity, while low-ranked items remain in obscurity.
We find that sensitivity to orders also differs across communities: some value a range of opinions, while others prefer a single authoritative answer.
Summary information displayed together can lead to \textit{presentation bias} \cite{Yue:2010}.
As the current voting scores are visibly presented with responses, users inevitably perceive the score before reading the contents of responses. 
Such exposure could immediately nudge user evaluations toward the majority opinion, making high\hyp{}scored responses more attractive.
We also find that the relative length of each response affects the polarity of future votes.

\begin{wraptable}[7]{r}{0.57\textwidth}
\vspace{-15px}
\setlength{\tabcolsep}{4pt}
\begin{center}
\small
\begin{tabular}{c | c c c | c }\toprule
\textbf{Res} & \textbf{Votes} & \textbf{Diff} & \textbf{Ratio} & \textbf{Relative Quality}\\ \hline
1 & $+ + + - - -$ & 0 & 0.5 & quite negative\\ 
2 & $+ - + - + -$ & 0 & 0.5 & moderately negative\\ 
3 & $- + - + - +$ & 0 & 0.5 & moderately positive\\ 
4 & $- - - + + +$ & 0 & 0.5 & quite positive\\ \bottomrule
\end{tabular}
\end{center}
\vspace{-5px}
\caption{Quality interpretation for each sequence of six votes.}
\label{tab:RelativeQuality}
\end{wraptable}

Standard discrete models for self\hyp{}reinforcing process include the Chinese Restaurant Process and the P{\'o}lya urn model.
Since these models are \textit{exchangeable}, the order of events does not affect the probability of a sequence.
However, Table \ref{tab:RelativeQuality} suggests how different \textit{contexts} of votes cause different impacts.
While the four sequences have equal numbers of positive and negative votes in aggregate, the fourth votes in the first and last responses are given against a clear majority opinion. 
Our model treats objection as a more challenging decision, thereby deserving higher weight.
In contrast, the middle two sequences receive alternating votes.
As each vote is a relatively weaker disagreement, the underlying quality is moderate compared to the other two responses.
Furthermore, if these are responses to one item, the order between them also matters.
If the initial three votes on the fourth response pushed its display position to the next page, for example, it might not have a chance to get future votes, which recover its reputation.

\begin{figure*}[h] 
\vspace{-5px}
\centering
\includegraphics[trim=0.0cm 0.0cm 0.0cm 0.0cm, width=0.99\textwidth]{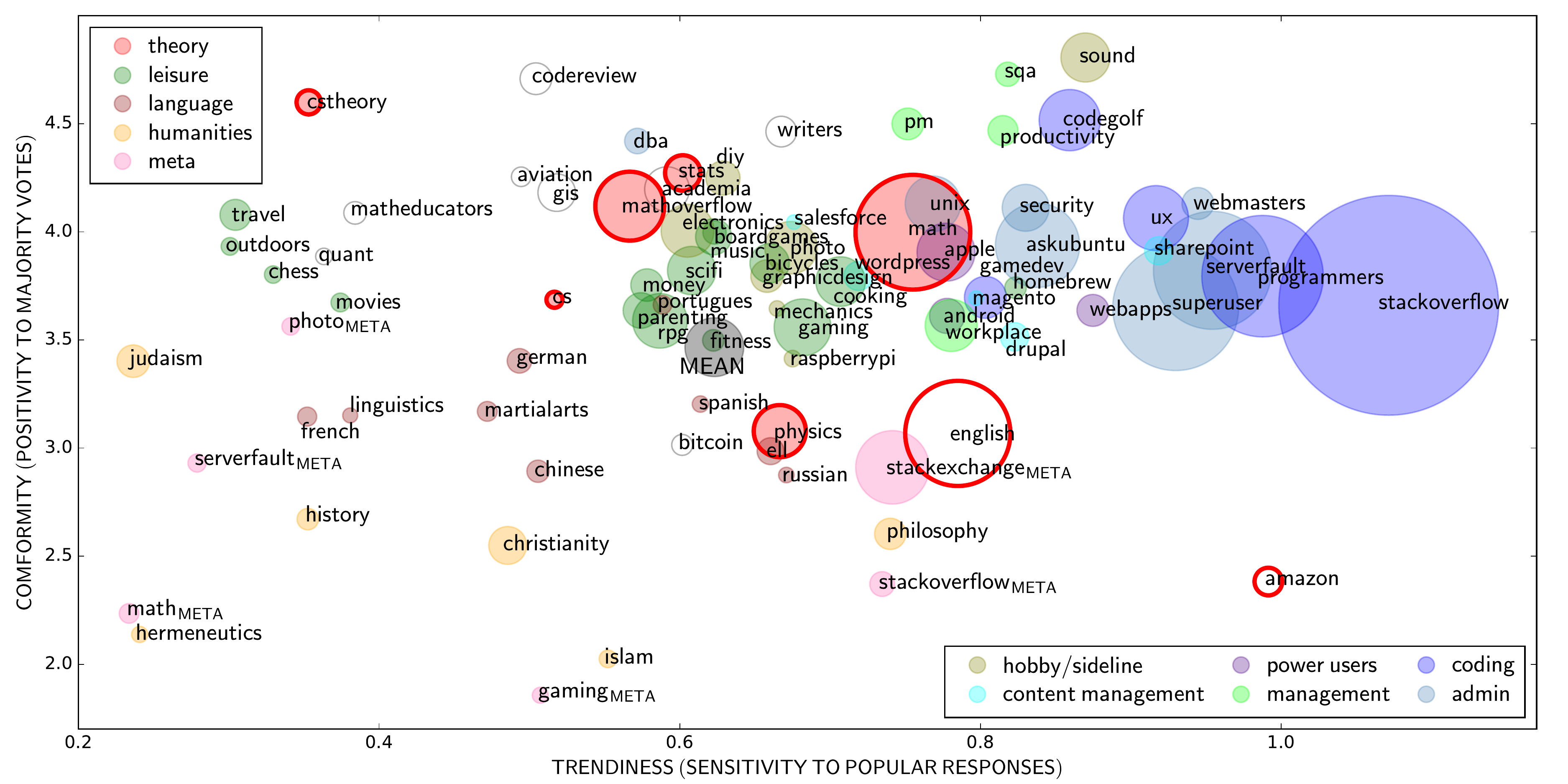}
\caption{2D Community embedding. Each of 83 communities is represented by two behavioral coefficients (\textit{Trendiness}, \textit{Conformity}). Eleven clusters are grouped based on their common focus. The MEAN community is synthesized by sampling 20 questions from every community (except Amazon due to the different user interface).}
\label{fig:Embedding}
\vspace{-15px}
\end{figure*}

The \textbf{Chinese Voting Process (CVP)} models generation of responses and votes, formalizing the evolution of helpfulness under \textit{positional} and \textit{presentational} reinforcement.
Whereas most previous work on helpfulness prediction \cite{Kim:2006, Ghose:2007, Liu:2007, Danescu:2009, Otterbacher:2009, Martin:2014, Siersdorfer:2014} has involved a single snapshot, the CVP estimates intrinsic quality of responses solely from selection and voting trajectories over multiple snapshots.
The resulting model shows significant improvements in predictive probability for helpfulness votes, especially in the critical early stages of a trajectory.
We find that the CVP estimated intrinsic quality ranks responses better than existing system rank, correlating orderly with the sentiment of comments associated with each response.
Finally, we qualitatively compare different characteristics of self\hyp{}reinforcing behavior between communities using two learned coefficients: \textit{Trendiness} and \textit{Conformity}.
The two-dimensional embedding in Figure \ref{fig:Embedding} characterizes different opinion dynamics from Judaism to Javascript (in StackOverflow).
\paragraph{Related work.}
\label{sec:related_work}
There is strong evidence that helpfulness voting is socially influenced.
Helpfulness ratings on Amazon product reviews differ significantly from independent human annotators \cite{Liu:2007}.
Votes are generally more positive, and the number of votes decreases exponentially based on displayed page position.
Review polarity is biased towards matching the consensus opinion \cite{Danescu:2009}:
when two reviews contain essentially the same text but differ in star rating, the review closer to the consensus star rating is considered more helpful.
There is also evidence that users vote strategically to correct perceived mismatches in review rank \cite{Sipos:2014}.
Many studies have attempted to predict helpfulness given review\hyp{}content features \cite{Kim:2006,Ghose:2007, Otterbacher:2009,Martin:2014,Siersdorfer:2014}.
Each of these examples predicts helpfulness based on text, star\hyp{}ratings, sales, and badges, but only at a single snapshot.
Our work differs in two ways. 
First, we combine data on Amazon helpfulness votes from \cite{Sipos:2014} with a much larger collection of helpfulness votes from 82 StackExchange forums.
Second, instead of considering text\hyp{}based features (which we hold out for evaluation) within a single snapshot, we attempt to predict the next vote at each stage based on the previous voting trajectory over multiple snapshots without considering textual contents.

%%%%%%%%%%%%%%%%%%%%%%%%%%%%%%%%%%%%%%%%%%%%%%%%%%%%%%%%%%%%%%%%%%
\section{The Chinese Voting Process}
  %%%%%%%%%%%%%%%%%%%%%%%%%%%%%
% 2. Chinese Voting Process %
%%%%%%%%%%%%%%%%%%%%%%%%%%%%%
\label{sec:model}

Our goal is to model helpfulness voting as a two-phase self-reinforcing stochastic process.
In the \textit{selection phase}, each user either selects an existing response based on their positions or writes a new response.
The positional reinforcement is inspired by the Chinese Restaurant Process (CRP) and Distance Dependent Chinese Restaurant Process (ddCRP). 
In the \textit{voting phase}, when one response is selected, the user chooses one of the two feedback options: a positive or negative vote based on the intrinsic quality and the presentational factors.
The presentational reinforcement is modeled by a log-linear model with time-varying features based on the P{\'o}lya urn model.
The CVP implements \textit{the\hyp{}rich\hyp{}get\hyp{}richer} dynamics as an interplay of these two preferential reinforcements, learning latent qualities of individual responses as inspired in Table \ref{tab:RelativeQuality}.
Specifically, each user at time $t$ interested in the item $i$ follows the generative story in Table \ref{tab:CVP}.
\setdefaultleftmargin{0cm}{1cm}{}{}{}{}
\begin{table}[ht]
\setlength{\tabcolsep}{3pt}
\small
\centering
\begin{tabular}{c | c}\toprule
  \textbf{Generative process} & \textbf{Sample parametrization (Amazon)}\\\hline
  \begin{minipage}[c]{0.57\textwidth}
    \begin{enumerate}
      \item Evaluate $j$-th response: $p(\MM{z}{i}{t}=j | \MM{z}{i}{1:t-1} ; \alpha) \propto \MM{f}{i}{t-1}(j)$
        \begin{enumerate}
          \item[(a)] `Yes': $p(\MM{v}{i}{t} = 1 | \bm{\theta}) = logit^{-1}(q_{ij} + \MM{g}{i}{t-1}(j))$
    	  \item[(b)] `No' : $p(\MM{v}{i}{t} = 0 | \bm{\theta}) = 1- p(\MM{v}{i}{t} = 1 | \bm{\theta})$
        \end{enumerate}
      \item Or write a new response: $p(\MM{z}{i}{t}=J_i+1 | \MM{z}{i}{1:t-1} ; \alpha) \propto \alpha$
        \begin{enumerate}
          \item[(a)] Sample $q_{i(J+1)}$ from $\mathcal{N}(0, \sigma^2).$
        \end{enumerate}
    \end{enumerate}
  \end{minipage}
  &%
  \begin{minipage}[c]{0.385\textwidth}
    \begin{equation*}
      \begin{aligned}
        \MM{f}{i}{t}(j) &= \Big(\frac{1}{1 +\MM{\textit{the-display-rank}}{i}{t}(j)}\Big)^{\tau} \\
        \MM{g}{i}{t}(j) &= \lambda \MM{r}{ij}{t} + \mu \MM{s}{ij}{t} + \nu_i \MM{u}{ij}{t} \\[3pt]
        \bm{\theta} &= \{\{q_{ij}\}, \lambda, \mu, \{\nu_i\}\}\\
        J_i &= \MM{J}{i}{t-1}  \; \text{(abbreviated notation)}\\[4pt]
      \end{aligned}
    \end{equation*}
  \end{minipage}\\\bottomrule
\end{tabular}
\caption{The generative story and the parametrization of the Chinese Voting Process (CVP).}
\label{tab:CVP}
\end{table}

%%%%%%%%%%%%%%%%%%%%%%%%%%%%%%%%%%%%%%%%%%%%%%%%%%%%%%%%%%%%%
\vspace{-20px}
\subsection{Selection phase} 
\label{sec:model:CRPs}
The CRP \cite{Aldous:1985, Blei:2003} is a self-reinforcing decision process over an infinite discrete set.
For each item (product/question) $i$, the first user writes a new response (review/answer).
The $t$-th subsequent user can choose an existing response $j$ out of $\M{J}{i}{t-1}$ possible responses with probability proportional to the number of votes $\M{n}{j}{t-1}$ given to the response $j$ by time $t-1$, whereas the probability of writing a new response $\M{J}{i}{t-1}+1$ is proportional to a constant $\alpha$.
While the CRP models self\hyp{}reinforcement --- each vote for a response makes that response more likely to be selected later --- there is evidence that the actual selection rate in an ordered list decays with display rank \cite{Joachims:2007}.
Since such rankings are mechanism-specific and not always clearly known in advance, we need a more flexible model that can specify various degrees of positional preference.
The ddCRP \cite{Blei:2011} introduces a function $f$ that decays with respect to some distance measure.
In our formulation, the distance function varies over time and is further configurable with respect to the specific interface of service providers.

Specifically, the function $\M{f}{i}{t}(j)$ in the CVP evaluates the popularity of the $j$-th response in the item $i$ at time $t$.
Since we assume that popularity of responses is decided by their positional accessibility, we can parametrize $f$ to be inversely proportional to their display ranks.
The exponent $\tau$ determines sensitivity to popularity in the selection phase by controlling the degree of harmonic penalization over ranks. 
Larger $\tau > 0$ indicates that users are more sensitive to trendy responses displayed near the top.
If $\tau < 0$, users often select low-ranked responses over high-ranked ones for some reasons.\footnote{This sometimes happens especially in the early stage when only a few responses exist.}
Note that even if the user at time $t$ does not vote on the $j$-th response, $\M{f}{i}{t}(j)$ could be different from $\M{f}{i}{t-1}(j)$ in the CVP,\footnote{Say the rank of another response $j'$ was lower than $j$'s at time $t-1$. If $t$-th vote given to the response $j'$ raises its rank higher than the rank of the response $j$, then $\M{f}{i}{t}(j)$ < $\M{f}{i}{t-1}(j)$ assuming $\tau > 0$.} whereas $\M{n}{ij}{t} = \M{n}{ij}{t-1}$ in the CRP.
Thus one can view the selection phase of the CVP as a \textit{non-exchangeable} extension of the CRP via a time-varying function $f$.

%%%%%%%%%%%%%%%%%%%%%%%%%%%%%%%%%%%%%%%%%%%%%%%%%%%%%%%%%%%%%
\subsection{Voting phase}
\label{sec:model:LR}

We next construct a self-reinforcing process for the inner voting phase.
The P{\'o}lya urn model is a self\hyp{}reinforcing decision process over a finite discrete set, but because it is exchangeable, it is unable to capture contextual information encoded in each a sequence of votes.
We instead use a log-linear formulation with the urn-based features, allowing other presentational features to be flexibly incorporated based on the modeler's observations.

Each response initially has $x=\MU{x}{0}$ positive and $y=\MU{y}{0}$ negative votes, which could be fractional pseudo-votes. 
For each draw of a vote, we return $w+1$ votes with the same polarity, thus self-reinforcing when $w>0$.
The following Table \ref{tab:PolyaUrn} shows time-evolving positive\slash{}negative ratios $\M{r}{j}{t}=\M{x}{j}{t}/(\M{x}{j}{t}+\M{y}{j}{t})$ and $\M{s}{j}{t}=\M{y}{j}{t}/(\M{x}{j}{t}+\M{y}{j}{t})$ of the first two responses: $j\in\{1,2\}$ in Table \ref{tab:RelativeQuality} with the corresponding ratio gain $\M{\Delta}{j}{t} = \M{r}{j}{t}-\M{r}{j}{t-1}$ (if $\M{v}{j}{t}=1$ or $+$) or $\M{s}{j}{t}-\M{s}{j}{t-1}$ (if $\M{v}{j}{t}=0$ or $-$).
\begin{table}[ht]
\setlength{\tabcolsep}{4pt}
\small
\centering
\begin{tabular}{c || c c c c | r || c c c c | r }\toprule
$t$ or $T$ & $\MM{v}{1}{t}$ & $\MM{r}{1}{t}$ & $\MM{s}{1}{t}$ & $\MM{\Delta}{1}{t}$ & \multicolumn{1}{c||}{$\MA{q}{1}{T}$} & $\MM{v}{2}{t}$ & $\MM{r}{2}{t}$ & $\MM{s}{2}{t}$ & $\MM{\Delta}{2}{t}$ & \multicolumn{1}{c}{$\MA{q}{2}{T}$} \\ \hline
0 &     & $1/2$ & $1/2$ &         &           &     & $1/2$ & $1/2$ &         &       \\
1 & $+$ & $2/3$ & $1/3$ & $0.167$ & \multicolumn{1}{c||}{$-$}        & $+$ & $2/3$ & $1/3$ & $0.167$ & \multicolumn{1}{c}{$-$}      \\
2 & $+$ & $3/4$ & $1/4$ & $0.083$ & 0.363     & $-$ & $2/4$ & $2/4$ & $0.167$ & -0.363      \\
3 & $+$ & $4/5$ & $1/5$ & $0.050$ & 0.574     & $+$ & $3/5$ & $2/5$ & $0.100$ & 0.004 \\
4 & $-$ & $4/6$ & $2/6$ & $0.133$ & 0.237     & $-$ & $3/6$ & $3/6$ & $0.100$ & -0.230 \\
5 & $-$ & $4/7$ & $3/7$ & $0.095$ & 0.004     & $+$ & $4/7$ & $3/7$ & $0.071$ & 0.007 \\
6 & $-$ & $4/8$ & $4/8$ & $0.071$ &{\bf -0.175}     & $-$ & $4/8$ & $4/8$ & $0.071$ & {\bf -0.166} \\ \bottomrule
\end{tabular}
\caption{Change of quality estimation $q_j$ over times for the first two example responses in Table \ref{tab:RelativeQuality} with the initial pseudo-votes $(x,y,w)=(1,1,1)$. The estimated quality at the first response sharply decreases when receiving the first majority-against vote at $t=4$. The first response ends up being more negative than the second, even if they receive the same number of votes in aggregate. These \textit{non-exchangeable} behaviors cannot be modeled with a simple exchangeable process.}
\label{tab:PolyaUrn}
\end{table}

In this toy setting, the polarity of a vote to a response is an outcome of its intrinsic quality as well as presentational factors: positive and negative votes.
Thus we model each sequence of votes by $\ell_2$-regularized logistic regression with the latent intrinsic quality and the P{\'o}lya urn ratios.\footnote{One might think \eqref{eqn:LR} can be equivalently achievable with only two parameters because of $\M{r}{j}{t} + \M{s}{j}{t}=1$ for all $t$. However, such reparametrization adds inconsistent translations to $\MA{q}{j}{T}$ and makes it difficult to interpret different inclinations between positive and negative votes for various communities.}
\begin{equation}
    \max_{\bm{\theta}} \;\; \log \prod_{t=2}^T logit^{-1}\big(\MA{q}{j}{T} + \lambda\M{r}{j}{t-1} + \mu\M{s}{j}{t-1}\big) - \frac{1}{2}\|\bm{\theta}\|_2^2 \quad \text{where } \; \bm{\theta}=\big(\MA{q}{j}{T}, \lambda, \mu\big)
\label{eqn:LR}
\end{equation}
The $\{\MA{q}{j}{T}\}$ in the Table \ref{tab:PolyaUrn} shows the result from solving \eqref{eqn:LR} up to $T$-th votes for each $j \in \{1, 2\}$.
The initial vote given at $t=1$ is disregarded in the training due to its arbitrariness from the uniform prior $(x_0 = y_0)$.
Since it is quite possible to have only positive or only negative votes, Gaussian regularization is necessary.
Note that using the urn-based ratio features is essential to encode contextual information.
If we instead use raw count features (only the numerators of $r_j$ and $s_j$), for example in the first response, the estimated quality $\MA{q}{1}{T}$ keeps increasing even after getting negative votes from time 4 to 6.
Log raw count features are unable to infer the negative quality.

In the first response, $\M{\Delta}{1}{t}$ shows the decreasing gain in positive ratios from $t=1$ to $3$ and in negative ratios from $t=4$ to $6$, whereas it gains a relatively large momentum at the first negative vote when $t=4$.
$\M{\Delta}{2}{t}$ converges to 0 in the 2nd response, implying that future votes have less effect than earlier votes for alternating $+/-$ votes.
$\MA{q}{2}{T}$ also converges to 0 as we expect neutral quality in the limit.
Overall the model is capable of learning intrinsic quality as desired in Table \ref{tab:RelativeQuality} where relative gains can be further controlled by tuning the initial pseudo-votes $(x, y)$.
   
In the real setting, the polarity score function $\M{g}{i}{t}(j)$ in the CVP evaluates presentational factors of the $j$-th response in the item $i$ at time $t$.
Because we adopt a log-linear formulation, one can easily add additional information about responses.
In addition to the positive ratio $\M{r}{ij}{t}$ and the negative ratio $\M{s}{ij}{t}$, $g$ also contains a length feature $\M{u}{ij}{t}$ (as given in Table \ref{tab:CVP}), which is the relative length of the response $j$ against the average length of responses in the item $i$ at particular time $t$.
Users in some items may prefer shorter responses than longer ones for brevity, whereas users in other items may blindly believe that longer responses are more credible before reading their contents.
The parameter $\nu_i$ explains length-wise preferential idiosyncrasy as a per-item bias: $\nu_i < 0$ means a preference toward the shorter responses.
Note that $\M{g}{i}{t}(j)$ could be different from $\M{g}{i}{t-1}(j)$ even if the user at time $t$ does not choose to vote.\footnote{If a new response is written at time $t$, $\M{u}{ij}{t} \neq \M{u}{ij}{t-1}$ as the new response changes the average length.}
All together, the voting phase of the CVP generates \textit{non-exchangeable} votes.

%%%%%%%%%%%%%%%%%%%%%%%%%%%%%%%%%%%%%%%%%%%%%%%%%%%%%%%%%%%%%%%%%%
\section{Inference}
  %%%%%%%%%%%%%%%%
% 3. Inference %
%%%%%%%%%%%%%%%%
\label{sec:inference}

Each phase of the CVP depends on the result of all previous stages, so decoupling these related problems is crucial for efficient inference.
We need to estimate community-level parameters, item-level length preferences, and response-level intrinsic qualities.
The graphical model of the CVP and corresponding parameters to estimate are illustrated in Table \ref{tab:GraphicalModel}.
We further compute two community-level behavioral coefficients: \textit{Trendiness} and \textit{Conformity}, which are useful summary statistics for exploring different voting patterns and explaining macro characteristics across different communities.

%%%%%%%%%%%%%%%%%%%%%%%%%%%%%%%%%%%%%%%%%%%%%%%%%%%%%%%%%%%%%
\paragraph{Parameter inference.}
The goal is to infer parameters $\bm{\theta}=\{\{q_{ij}\}, \lambda, \mu, \{\nu_i\}\}$. 
We sometimes use $f$ and $g$ instead to compactly indicate parameters associated to each function.
The likelihood of one CVP step in the item $i$ at time $t$ is $L_i^{(t)}(\tau, \bm{\theta}; \alpha, \sigma) =$
\begin{Equation*}{0.93} \label{cvp:Likelihood}
\Big\{\frac{\alpha}{\alpha + \sum_{j=1}^{\MM{J}{i}{t-1}}\MM{f}{i}{t-1}(j)} \mathcal{N}(q_{i,\MM{z}{i}{t}} ; 0, \sigma^2)\Big\}^{\mathbbm{1}[\MM{z}{i}{t} = \MM{J}{i}{t-1} + 1]} 
\Big\{ \frac{\MM{f}{i}{t-1}(\MM{z}{i}{t}) }{ \alpha + \sum_{j=1}^{\MM{J}{i}{t-1}}\MM{f}{i}{t-1}(j) } p(\MM{v}{i}{t} | q_{i, \MM{z}{i}{t}}, \MM{g}{i}{t-1}(j)) \Big\}^{\mathbbm{1}[\MM{z}{i}{t} \leq \M{J}{i}{t-1}]}
\end{Equation*}
where the two terms correspond to writing a new response and selecting an existing response to vote.
The fractions in each term respectively indicate the probability of writing a new response and choosing existing responses in the selection phase.
The other two probability expression in each term describe quality sampling from a normal distribution and the logistic regression in the voting phase.

\begin{table}[t]
\setlength{\tabcolsep}{3pt}
\small
\centering
\begin{tabular}{c p{0.05cm} c}\\
  \begin{minipage}[c]{0.47\textwidth}
    \includegraphics[trim=0.5cm 0.5cm 0.5cm 0.5cm, width=1.0\columnwidth]{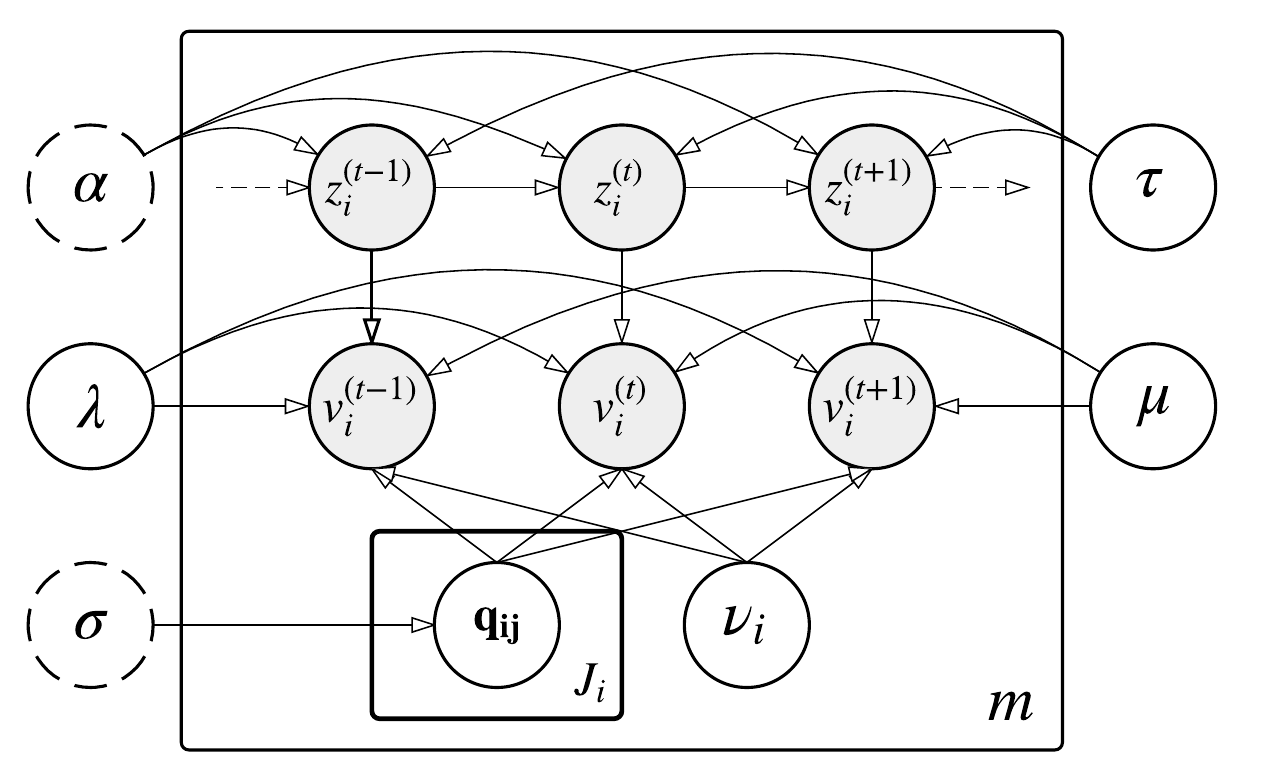}
  \end{minipage}
  &%
  &%
  \begin{minipage}[c]{0.48\textwidth}
    \begin{itemize}
      \item[] $\alpha$:   hyper-parameter for response growth
      \item[] $\sigma^2$: hyper-parameter for quality variance
      \item[] $\tau$:     community-level sensitivity to popularity
      \item[] $\lambda$:  community-level preference for positive ratio
      \item[] $\mu$:      community-level preference for negative ratio
      \item[] $\nu_i$:    item-level preference for response length
      \item[] $q_{ij}$:   response-level hidden intrinsic quality 
      \item[] $m$:        \# of items (e.g., products/questions)
      \item[] $J_i$:      \# of responses of item $i$ (e.g., reviews/answers)
    \end{itemize}
  \end{minipage}\\\\
\end{tabular}
\caption{Graphical model and parameters for the CVP. Only three time steps are unrolled for visualization.}
\label{tab:GraphicalModel}
\end{table}
It is important to note that our setting differs from many CRP-based models.
The CRP is typically used to represent a non-parametric prior over the choice of latent cluster assignments that must themselves be inferred from noisy observations.
In our case, the result of each choice is directly observable because we have the \textit{complete trajectory} of helpfulness votes.
As a result, we only need to infer the continuous parameters of the process, and not combinatorial configurations of discrete variables.
Since we know the complete trajectory where the rank inside the function $f$ is a part of the true observations, we can view each vote as an independent sample.
Denoting the last timestamp of the item $i$ by $T_i$, the log-likelihood becomes $\smash{\ell(\tau, \bm{\theta}; \alpha, \sigma) = \sum_{i=1}^m \sum_{t=1}^{T_i} \log{L_i^{(t)}}}$ and is further separated into two pieces:
\begin{align}
    \ell_{v}(\bm{\theta} ; \sigma) &= \sum_{i=1}^m \sum_{t=1}^{T_i} \Big\{\mathbbm{1}[write] \cdot \log \mathcal{N}(q_{i,\MM{z}{i}{t}} ; 0, \sigma^2) + {\mathbbm{1}[choose]} \cdot \log p(\MM{v}{i}{t} | q_{i, \MM{z}{i}{t}}, \MM{g}{i}{t-1}(j)) \Big\},\\
     \ell_{s}(\tau ; \alpha) &= \sum_{i=1}^m \sum_{t=1}^{T_i} \Big\{\mathbbm{1}[write] \cdot \log \frac{\alpha}{\alpha + \sum_{j=1}^{\MM{J}{i}{t-1}}\MM{f}{i}{t-1}(j)} + {\mathbbm{1}[choose]} \cdot \log \frac{\MM{f}{i}{t-1}(\MM{z}{i}{t}) }{ \alpha + \sum_{j=1}^{\MM{J}{i}{t-1}}\MM{f}{i}{t-1}(j)} \Big\}.\nonumber
\end{align}
Inferring a whole trajectory based only on the final snapshots would likely be intractable for a non-exchangeable model.
Due to the continuous interaction between $f$ and $g$ for every time step, small mis-predictions in the earlier stages will cause entirely different configurations.
Moreover the rank function inside $f$ is in many cases site-specific.\footnote{We generally know that Amazon decides the display order by the portion of positive votes and the total number of votes on each response, but the relative weights between them are not known. We do not know how StackExchange forums break ties, which affects highly in the early stages of voting.}
It is therefore vital to observe all trajectories of random variables $\{\M{z}{i}{t}, \M{v}{i}{t}\}$: decoupling $f$ and $g$ reduces the inference problem into estimating parameters separately for the selection phase and the voting phase.
Maximizing $\ell_{v}$ can be efficiently solved by $\ell_2$-regularized logistic regression as demonstrated for \eqref{eqn:LR}.
If the hyper-parameter $\alpha$ is fixed, maximizing $\ell_{s}$ becomes a convex optimization because $\tau$ appears in both the numerator and the denominator.
Since the gradient for each parameter in $\bm{\theta}$ is obvious, we only include the gradient of $\M{\ell}{s,i}{t}$ for the particular item $i$ at time $t$ with respect to $\tau$.
Then $\frac{\partial \ell_s}{\partial \tau} = \sum_{i=1}^m \sum_{t=1}^{T_i} {\partial \ell_{s,i}^{(t)}}/{\partial \tau}$.
\begin{align}
\frac{\partial \ell_{s,i}^{(t)}}{\partial \tau} 
= \frac{1}{\tau} \Bigg\{ \mathbbm{1}[\MM{z}{i}{t} \le \MM{J}{i}{t-1}] \cdot
 \frac{\MM{f}{i}{t-1}(\MM{z}{i}{t}) \log \MM{f}{i}{t-1}(\MM{z}{i}{t})}{\MM{f}{i}{t-1}(\MM{z}{i}{t})} 
- \frac{\sum_{j=1}^{\MM{J}{i}{t-1}} \MM{f}{i}{t-1}(j) \log \MM{f}{if}{t-1}(j)}{\alpha + \sum_{j = 1}^{\MM{J}{i}{t-1}} \MM{f}{i}{t-1}(j)}  \Bigg\}
\end{align}

%%%%%%%%%%%%%%%%%%%%%%%%%%%%%%%%%%%%%%%%%%%%%%%%%%%%%%%%%%%%%
\paragraph{Behavioral coefficients.}
To succinctly measure overall voting behaviors across different communities, we propose two community-level coefficients.
\textit{Trendiness} indicates the sensitivity to positional popularity in the selection phase.
While the community-level $\tau$ parameter renders Trendiness simply to avoid overly-complicated models, one can easily extend the CVP to have per-item $\tau_i$ to better fit the data. 
In that case, Trendiness would be a summary statistics for $\{\tau_i\}$.
\textit{Conformity} captures users' receptiveness to prevailing polarity in the voting phase.
To count every single vote, we define Conformity to be a geometric mean of odds ratios between majority-following votes and majority-disagreeing votes.
Let $V_i$ be the set of time steps when users vote rather than writing responses in the item $i$.
Say $n$ is the total number of votes across all items in the target community. 
Then Conformity is defined as
\begin{equation*}
  \kappa = \Bigg\{\prod_{i=1}^m\prod_{t \in V_i} \bigg(\frac{P(\MM{v}{i}{t+1}=1 | q_{i, \MM{z}{i}{t+1}}^{t}, \lambda^{t}, \mu^{t}, \nu_i^{t})}{P(\MM{v}{i}{t+1}=0 | q_{i, \MM{z}{i}{t+1}}^{t}, \lambda^{t}, \mu^{t}, \nu_i^{t})}\bigg)^{\MM{h}{i}{t}} \Bigg\}^{1/n} \quad \text{where} \;\; \MM{h}{i}{t} = \begin{cases}1 & (\MP{n}{ij}{t} \geq \MN{n}{ij}{t})\\-1 & (\MP{n}{ij}{t} < \MN{n}{ij}{t}) \end{cases}.
\end{equation*}
To compute Conformity $\kappa$, we need to learn $\bm{\theta}^t = \{q_{ij}^t, \lambda^t, \mu^t, \nu_i^t\}$ for each $t$, which is a set of parameters learned on the data only up to the time $t$.
This is because the user at time $t$ cannot see any future which will be given later than the time $t$.
Note that $\bm{\theta}^{t+1}$ can be efficiently learned  by \textit{warm-starting} at $\bm{\theta}^t$.
In addition, while positive votes are mostly dominant in the end, the dominant mood up to time $t$ could be negative, exactly when the user at time $t+1$ tries to vote.
In this case, $\M{h}{i}{t}$ becomes $-1$, inverting the fraction to be the ratio of following the majority against the minority.
By summarizing learned parameters in terms of two coefficients $(\tau, \kappa)$, we can compare different selection\slash{}voting behaviors for various communities.

%%%%%%%%%%%%%%%%%%%%%%%%%%%%%%%%%%%%%%%%%%%%%%%%%%%%%%%%%%%%%%%%%%
\section{Experiments}
  %%%%%%%%%%%%%%%%%%
% 4. Experiments %
%%%%%%%%%%%%%%%%%%
\label{sec:experiments}

\begin{table}[t]
\setlength{\tabcolsep}{3pt}
\small
\centering
\begin{tabular}{l || c c | c c c c c c c | c c | c c }\toprule
\multirow{2}{*}{\textbf{Community}} & \multicolumn{2}{c|}{\textbf{Selection}} & \multicolumn{7}{c|}{\textbf{Voting}} & \multicolumn{2}{c|}{\textbf{Residual}} & \multicolumn{2}{c}{\textbf{Bumpiness}}\\ \cline{2-14}
 & \textit{CRP} & \textit{CVP} & $q_{ij}$ & $\lambda$ & $\nu_i$ & $q_{ij}, \lambda$ & $q_{ij}, \nu_i$ & $\lambda, \nu_i$ & \textit{Full} & \textit{Rank} & \textit{Qual} & \textit{Rank} & \textit{Qual}\\ \hline
SOF$_{(22925)}$ & 2.152 & \textbf{{\color{sig01}1.989}} & {\color{sig001}.107} & {\color{sig001}.103} & {\color{sig001}.108} & {\color{sig001}.100} & {\color{sig001}.106} & {\color{sig001}.100} & \textbf{.096} & .005 & \textbf{.003} & .080 & \textbf{.038}\\
math$_{(6245)}$ & \textbf{1.841} & 1.876 & {\color{sig001}.071} & {\color{sig001}.064} & {\color{sig001}.067} & {\color{sig001}.062} & {\color{sig001}.066} & {\color{sig001}.060} & \textbf{.059} & .014 & \textbf{.008} & .280 & \textbf{.139}\\
english$_{(5242)}$ & 1.969 & \textbf{1.924} & {\color{sig001}.160} & {\color{sig001}.146} & {\color{sig001}.152} & {\color{sig001}.141} & {\color{sig001}.147} & {\color{sig001}.137} & \textbf{.135} & .018 & \textbf{.007} & .285 & \textbf{.149}\\
mathOF$_{(2255)}$ & 1.992 & \textbf{{\color{sig01}1.910}} & {\color{sig001}.049} & {\color{sig001}.046} & {\color{sig001}.049} & {\color{sig01}.045} & {\color{sig001}.047} & {\color{sig001}.046} & \textbf{.045} & .009 & \textbf{.007} & .185 & \textbf{.119}\\
physics$_{(1288)}$ & 1.824 & \textbf{1.801} & {\color{sig001}.174} & {\color{sig001}.155} & {\color{sig001}.166} & {\color{sig001}.150} & {\color{sig001}.156} & {\color{sig001}.146} & \textbf{.142} & .032 & \textbf{.014} & .497 & \textbf{.273}\\
stats$_{(598)}$ & 1.889 & \textbf{{\color{sig001}1.822}} & {\color{sig001}.051} & {\color{sig001}.044} & {\color{sig001}.048} & {\color{sig001}.043} & {\color{sig001}.046} & {\color{sig001}.042} & \textbf{.042} & .030 & \textbf{.019} & .613 & \textbf{.347}\\
judaism$_{(504)}$ & 2.039 & \textbf{{\color{sig001}1.859}} & {\color{sig001}.135} & {\color{sig001}.124} & {\color{sig001}.132} & {\color{sig001}.121} & {\color{sig001}.125} & {\color{sig01}.118} & \textbf{.116} & .046 & \textbf{.018} & .875 & \textbf{.403}\\
amazon$_{(363)}$ & 2.597 & \textbf{{\color{sig001}2.261}} & {\color{sig001}.266} & {\color{sig001}.270} & {\color{sig001}.262} & {\color{sig001}.254} & {\color{sig001}.243} & {\color{sig001}.253} & \textbf{.240} & .023 & \textbf{.016} & .392 & \textbf{.345}\\
meta.SOF$_{(294)}$ & \textbf{{\color{sig001}1.411}} & 1.575 & {\color{sig001}.261} & {\color{sig001}.241} & {\color{sig001}.270} & {\color{sig001}.229} & {\color{sig001}.243} & {\color{sig001}.232} & \textbf{.225} & .018 & \textbf{.013} & .281 & \textbf{.255}\\
cstheory$_{(279)}$ & 1.893 & \textbf{{\color{sig001}1.795}} & {\color{sig001}.052} & {\color{sig001}.040} & {\color{sig001}.053} & {\color{sig01}.039} & {\color{sig001}.049} & {\color{sig001}.039} & \textbf{.038} & .032 & \textbf{.029} & \textbf{.485} & .553\\
cs$_{(123)}$ & 1.825 & \textbf{1.780} & {\color{sig001}.128} & {\color{sig05}.100} & {\color{sig001}.118} & {\color{sig05}.099} & {\color{sig001}.113} & .097 & \textbf{.096} & .069 & \textbf{.040} & .725 & \textbf{.673}\\
linguistics$_{(107)}$ & 1.993 & \textbf{{\color{sig001}1.789}} & {\color{sig001}.133} & {\color{sig001}.127} & {\color{sig001}.130} & {\color{sig001}.122} & {\color{sig001}.123} & {\color{sig001}.120} & \textbf{.116} & .074 & \textbf{.038} & .778 & \textbf{.656}\\ \hline
AVERAGE & 2.050 & \textbf{1.945} & .109 & .103 & .108 & .099 & .105 & .098 & \textbf{.095} & .011 & \textbf{.006} & .186 & \textbf{.101}\\
\bottomrule
\end{tabular}
\caption{Predictive analysis on the first 50 votes: In the selection phase, the CVP shows better negative log-likelihood in almost all forums. In the voting phase, the full model shows better negative log-likelihood than all subsets of features. Quality analysis at the final snapshot: Smaller residuals and bumpiness show that the order based on the estimated quality $q_{ij}$ more coherently correlates with the average sentiments of the associated comments than the order by display rank. (SOF=StackOverflow, OF=Overflow, rest=Exchange, {\color{sig001}Blue}: $p \leq 0.001$, {\color{sig01}Green}: $p \leq 0.01$, {\color{sig05}Red}: $p \leq 0.05$)}
\label{tab:Results}
\vspace{-15px}
\end{table}

We evaluate the CVP on product reviews from Amazon and 82 issue-specific forums from the StackExchange network.
The Amazon dataset  \cite{Sipos:2014} originally consisted of 595 products with daily snapshots of writing\slash{}voting trajectories from Oct 2012 to Mar 2013.
After eliminating duplicate products\footnote{Different seasons of the same TV shows have different ASIN codes but share the same reviews.} and products with fewer than five reviews or fragmented trajectories,\footnote{If the number of total votes between the last snapshot of the early fragment and the first snapshot of the later fragment is less than 3, we fill in the missing information simply with the last snapshot of the earlier fragment.} 363 products are left. 
For the StackExchange dataset\footnote{Dataset and statistics are available at \url{https://archive.org/details/stackexchange}.}, we filter out questions from each community with fewer than five answers besides the answer chosen by the question owner.\footnote{The answer selected by the question owner is displayed first regardless of voting scores.}
We drop communities with fewer than 100 questions after pre-processing.
Many of these are ``Meta'' forums where users discuss policies and logistics for their original forums.

\begin{figure}[t]
    \centering
    \includegraphics[trim={5.0cm 1.5cm 5.0cm 2.0cm},clip, width=0.95\columnwidth]{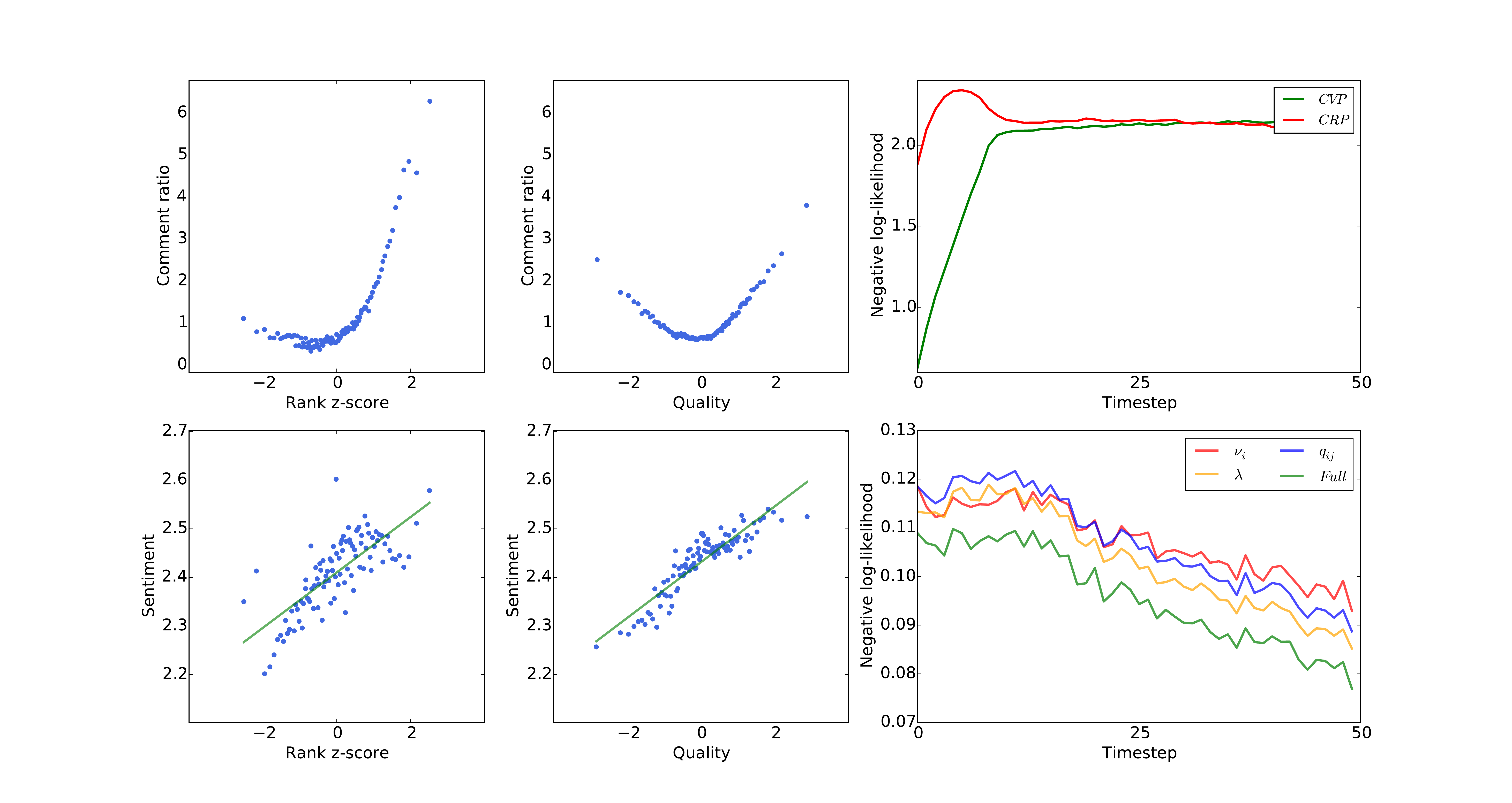}
    \caption{Comment and likelihood analysis on the StackOverflow forum. The left panels show that responses with higher ranks tend to have more comments (top) and more positive sentiments (bottom). The middle panels show responses have more comments at both high and low intrinsic quality $q_{ij}$ (top). The corresponding sentiment correlates more cohesively with the quality score (bottom). Each blue dot is approximately an average over 1k responses, and we parse 337k comments given on 104k responses in total. The right panels show predictive power for the selection phase (top) and the voting phase (bottom) up to $t < 50$ (lower is better).}
    \label{fig:comments}       
    \vspace{-10px}
\end{figure}

\paragraph{Predictive analysis.} 
In each community, our prediction task is to learn the model up to time $t$ and predict the action at $t+1$.
We align all items at their initial time steps and compute the average negative log-likelihood of the next actions based on the current model.
Since the complete trajectory enables us to separate the selection and voting phases in inference, we also measure the predictive power of these two tasks separately against their own baselines.
For the selection phase, the baseline is the CRP, which selects responses proportional to the number of accumulated votes or writes a new response with the probability proportional to $\alpha$.\footnote{We fix $\alpha$ to 0.5 after searching over a wide range of values.}
When $t<50$, as shown in the first column of Table \ref{tab:Results}, the CVP significantly outperforms the CRP based on paired $t$-tests (two-tailed).
Using the function $f$ based on display rank and Trendiness parameter $\tau$ is indeed a more precise representation of positional accessibility.
Especially in the early stages, users often select responses displayed at lower ranks with fewer votes.
While the CRP has no ability to give high scores in these cases, the CVP properly models it by decreasing $\tau$.
The comparative advantage of the CVP declines as more votes become available and the correlation between display rank and the number of votes increases. 
For items with $t \geq 50$, there is no significant difference between the two models as exemplified in the third column of Figure \ref{fig:comments}.
These results are coherent across other communities ($p > 0.07$).

Improving predictive power on the voting phase is difficult because positive votes dominate in every community.
We compare the fully parametrized model to simpler partial models in which certain parameters are set to zero.
For example, a model with all parameters but $\lambda$ knocked out is comparable to a plain P{\'o}lya Urn.
As illustrated in the second column of Table \ref{tab:Results}, we verify that every sub-model is significantly different from the full model in all major communities based on one-way ANOVA test, implying that each feature adds distinctive and meaningful information.
Having the item-specific length bias $\nu_i$ provides significant improvements as well as having intrinsic quality $q_{ij}$ and current opinion counts $\lambda$. 
While we omit the log-likelihood results with $t \geq 50$, all model better predicts true polarity when $t \geq 50$, because the log-linear model obtains a more robust estimate of community-level parameters as the model acquires more training samples.

\vspace{-5px}
\paragraph{Quality analysis.}
The primary advantage of the CVP is its ability to learn ``intrinsic quality'' for each response that filters out noise from self-reinforcing voting processes.
We validate these scores by comparing them to another source of user feedback: 
both StackExchange and Amazon allow users to attach comments to responses along with votes.
For each response, we record the number of comments and the average sentiment of those comments as estimated by \cite{Socher:2013}.
As a baseline, we also calculate the final display rank of each response, which we convert to a z-score to make it more comparable to the quality scores $q_{ij}$.
After sorting responses based on display rank and quality rank, we measure the association between the two rankings and comment sentiment with linear regression.
Results are shown for StackOverflow in Figure \ref{fig:comments}.
As expected, highly-ranked responses have more comments, but we also find that there are more comments for both high {\em and} low values of intrinsic quality.
Both better display rank and higher quality score $q_{ij}$ are clearly associated with more positive comments (slope $\in$ [0.47, 0.64]), but the residuals of quality rank 0.012 are on average less than the half the residuals of display rank 0.028.
In addition, we also calculate the ``bumpiness'' of these plots by computing the mean variation of two consecutive slopes between each adjacent pair of data points.
Quality rank reduces bumpiness of display rank from 0.391 to 0.226 in average, implying the estimated intrinsic quality yields locally consistent ranking as well as globally consistent.\footnote{All numbers and p-values in paragraphs are weighted averages on all 83 communities, whereas Table \ref{tab:Results} only includes results for the major communities and their own weighted averages due to space limits.}

\begin{wrapfigure}{R}{7.5cm}
\centering
    \includegraphics[trim=1.5cm 0.2cm 0.5cm 0.5cm, width=0.45\columnwidth]{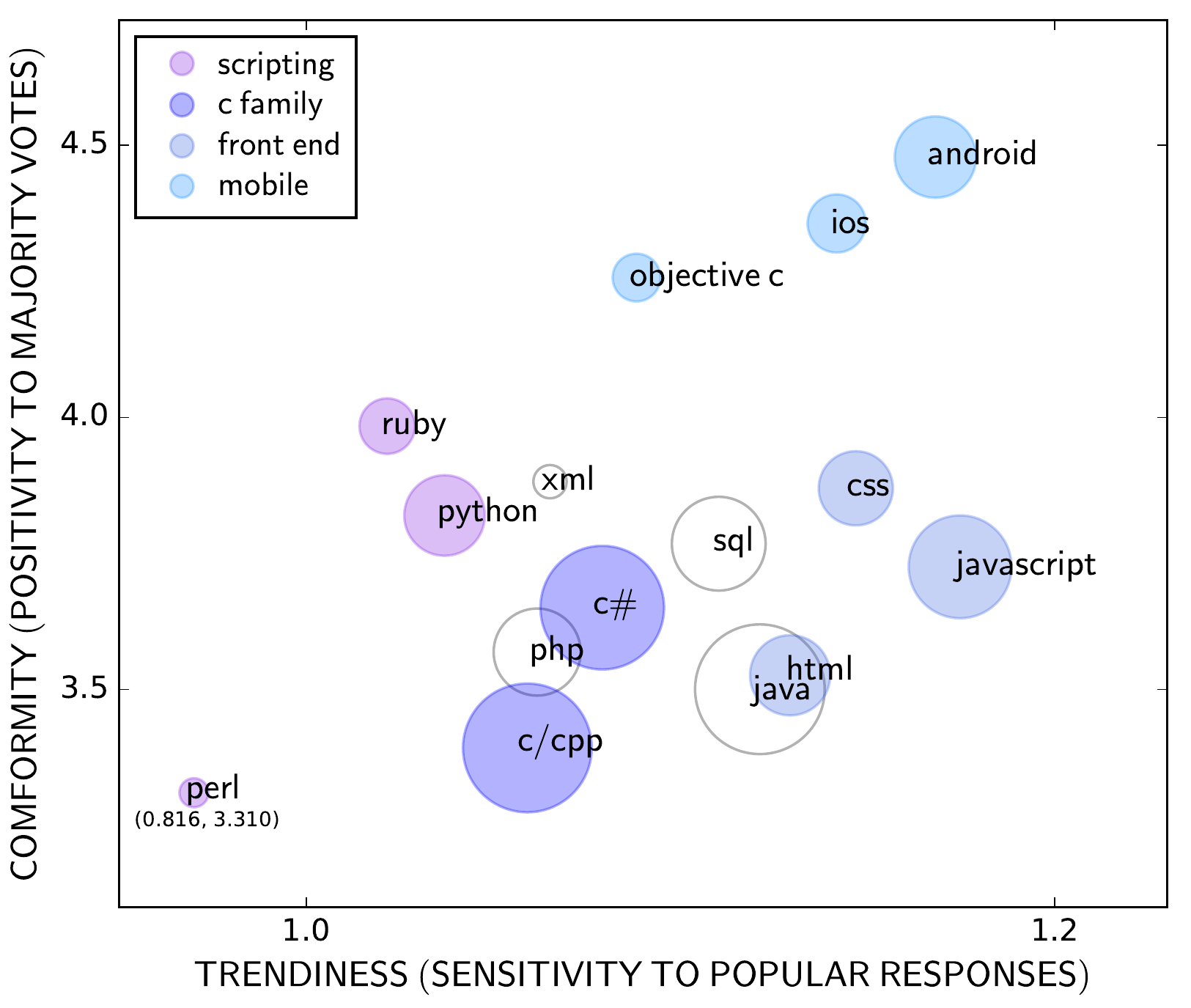}
    \caption{Sub-community embedding for StackOverflow.}
    \label{fig:EmbeddingSub}
    \vspace{-10px}
\end{wrapfigure}

\vspace{-5px}
\paragraph{Community analysis.}
The 2D embedding in Figure \ref{fig:Embedding} shows that we can compare and contrast the different evaluation cultures of communities using two inferred behavioral coefficients: Trendiness $\tau$ and Conformity $\kappa$. 
Communities are sized according to the number of items and colored based on a manual clustering.
Related communities collocate in the same neighborhood. 
Religion, scholarship, and meta-discussions cluster towards the bottom left, where users are interested in many different opinions, and are happy to disagree with each other.
Going from left to right, communities become more trendy:
users in trendier communities tend to select and vote mostly on already highly-ranked responses.
Going from bottom to top, users become increasingly likely to conform to the majority opinion on any given response.
By comparing related communities we can observe that characteristics of user communities determine voting behavior more than technical similarity.
Highly theoretical and abstract communities (\textit{cstheory}) have low Trendiness but high Conformity.
More applied, but still graduate-level, communities in similar fields (\textit{cs, mathoverflow, stats}) show less Conformity but greater Trendiness.
Finally, more practical homework-oriented forums (\textit{physics, math}) are even more trendy.
In contrast, users in \textit{english} are trendy and debatable.
Users in \textit{Amazon} are most sensitive to trendy reviews and least afraid of voicing minority opinion.

StackOverflow is by far the largest community, and it is reasonable to wonder whether the Trendiness parameter is simply a proxy for size.
When we subdivide StackOverflow by programming languages however (see Figure \ref{fig:EmbeddingSub}), individual community averages can be distinguished, but they all remain in the same region.
\textit{Javascript} programmers are more satisfied with trendy responses than those using \textit{c/c++}.
Mobile developers tend to be more conformist, while \textit{Perl} hackers are more likely to argue.

%%%%%%%%%%%%%%%%%%%%%%%%%%%%%%%%%%%%%%%%%%%%%%%%%%%%%%%%%%%%%%%%%%
\section{Conclusions}
  Helpfulness voting is a powerful tool to evaluate user-generated responses such as product reviews and question answers.
However such votes can be socially reinforced by positional accessibility and existing evaluations by other users.
In contrast to many exchangeable random processes, the CVP takes into account sequences of votes, assigning different weights based on the \textit{context that each vote was cast}.
Instead of trying to model the response ordering function $f$, which is mechanism-specific and often changes based on service providers' strategies, we leverage the fully observed trajectories of votes, estimating the hidden intrinsic quality of each response and inferring two behavioral coefficients for community-level exploration.
The proposed log-linear urn model is capable of generating non-exchangeable votes with great scalability to incorporate other factors such as length bias or other textual features. 
As we are more able to observe social interactions \textit{as they are occurring} and not just summarized after the fact, we will increasingly be able to use models beyond exchangeability.

%%%%%%%%%%%%%%%%%%%%%%%%%%%%%%%%%%%%%%%%%%%%%%%%%%%%%%%%%%%%%%%%%%
% \subsubsection*{Acknowledgments}

% Use unnumbered third level headings for the acknowledgments. All
% acknowledgments go at the end of the paper. Do not include
% acknowledgments in the anonymized submission, only in the final paper.

%%%%%%%%%%%%%%%%%%%%%%%%%%%%%%%%%%%%%%%%%%%%%%%%%%%%%%%%%%%%%%%%%%
\bibliographystyle{abbrv}
\small{
\bibliography{references}
}

\end{document}